\documentclass[%
 aip,
 amsmath,amssymb,
 reprint,%
]{revtex4-1}

\usepackage{graphicx}
\usepackage{dcolumn}
\usepackage{bm}

\usepackage[utf8]{inputenc}
\usepackage[T1]{fontenc}
\usepackage{mathptmx}
\usepackage{hyperref}
\usepackage{color}
\usepackage{epsfig} 
\hypersetup{colorlinks=true, citecolor = black, urlcolor=blue, }
\begin{document}

\preprint{AIP/123-QED}

\title{Deep learning-based statistical noise reduction for multidimensional spectral data}

\author{Younsik Kim}
\author{Dongjin Oh}
\author{Soonsang Huh}
\author{Dongjoon Song}
\affiliation{Center for Correlated Electron Systems, Institute for Basic Science, Seoul, 08826, Korea}
\affiliation{Department of Physics and Astronomy, Seoul National University, Seoul, 08826, Korea}

\author{Sunbeom Jeong}
\affiliation{Department of Electrical and Computer Engineering, Seoul National University, Seoul, 08826, Korea}

\author{Junyoung Kwon}
\author{Minsoo Kim}
\author{Donghan Kim}
\author{Hanyoung Ryu}
\author{Jongkeun Jung}
\author{Wonshik Kyung}
\author{Byungmin Sohn}
\author{Suyoung Lee}
\affiliation{Center for Correlated Electron Systems, Institute for Basic Science, Seoul, 08826, Korea}
\affiliation{Department of Physics and Astronomy, Seoul National University, Seoul, 08826, Korea}

\author{Jounghoon Hyun}
\author{Yeonghoon Lee}
\author{Yeongkwan Kim}
\affiliation{Department of Physics, Korea Advanced Institute of Science and Technology, Daejeon, 34141, Korea}
\author{Changyoung Kim}
\email[Electronic address:$~$]{changyoung@snu.ac.kr}
\affiliation{Center for Correlated Electron Systems, Institute for Basic Science, Seoul, 08826, Korea}
\affiliation{Department of Physics and Astronomy, Seoul National University, Seoul, 08826, Korea}

\date{\today}

\begin{abstract}
In spectroscopic experiments, data acquisition in multi-dimensional phase space may require long acquisition time, owing to the large phase space volume to be covered. In such case, the limited time available for data acquisition can be a serious constraint for experiments in which multidimensional spectral data are acquired. Here, taking angle-resolved photoemission spectroscopy (ARPES) as an example, we demonstrate a denoising method that utilizes deep learning as an intelligent way to overcome the constraint. With readily available ARPES data and random generation of training data set, we successfully trained the denoising neural network without overfitting. The denoising neural network can remove the noise in the data while preserving its intrinsic information. We show that the denoising neural network allows us to perform similar level of second-derivative and line shape analysis on data taken with two orders of magnitude less acquisition time. The importance of our method lies in its applicability to any multidimensional spectral data that are susceptible to statistical noise.
\end{abstract}

\maketitle

\begin{figure*}[htbp]
\includegraphics[width=1\textwidth]{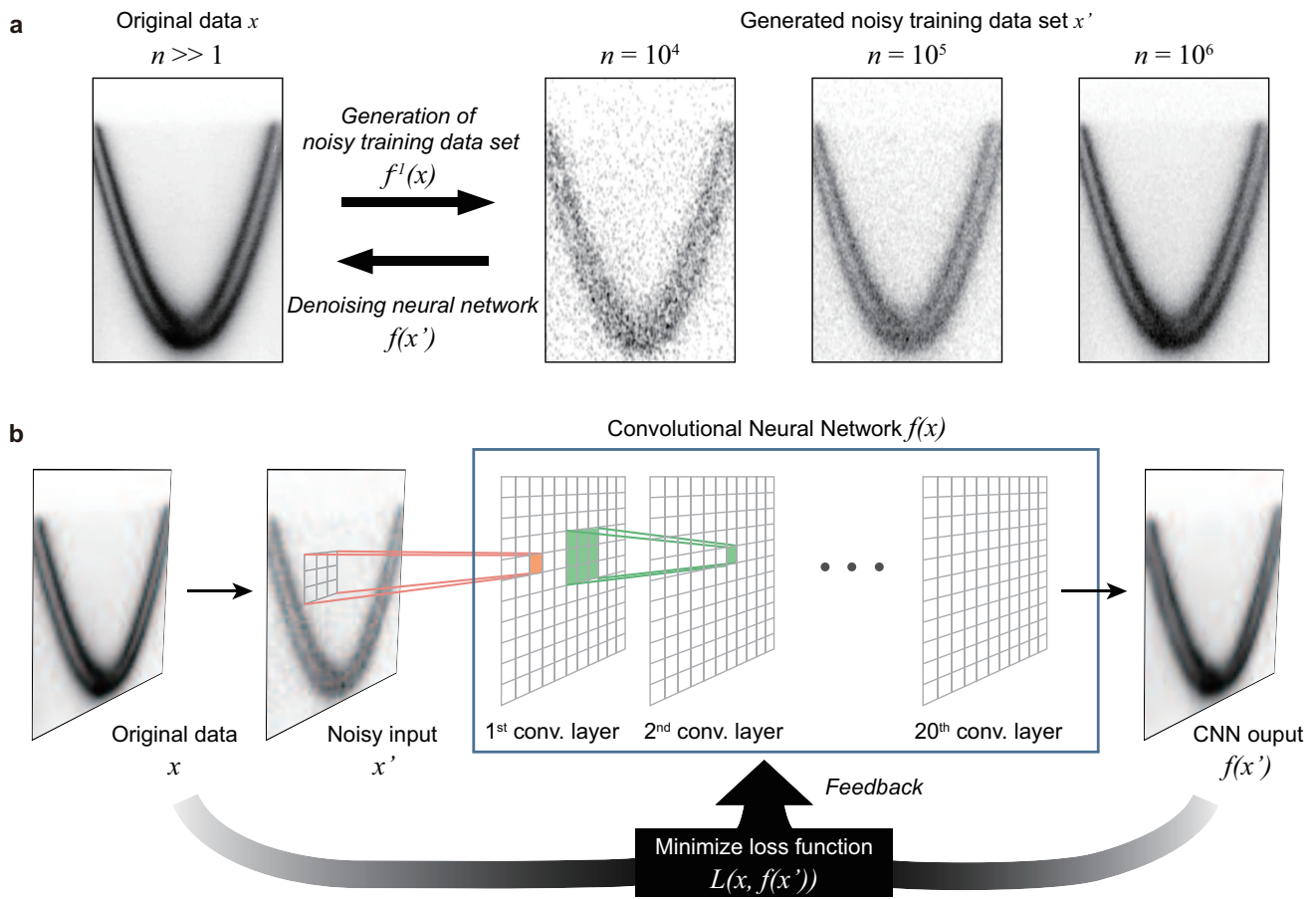}
\caption{Overall training sequence of denoising neural network. (a) Schematic for the generation of noisy training data set. $n$ denotes the number of counts in the spectrum. $x$ denotes original data and $x'$ denotes  noisy data generated from original data $x$. The denoising process is denoted as $f(x)$, while the generation of noisy training data set which is the inverse-process of denoising is denoted as $f^{-1}(x')$. The data used here is Au(111) surface state ARPES data. (b) Schematic of training the denoising neural network.}

\end{figure*}

\begin{figure*}[htbp]
\includegraphics[scale=1]{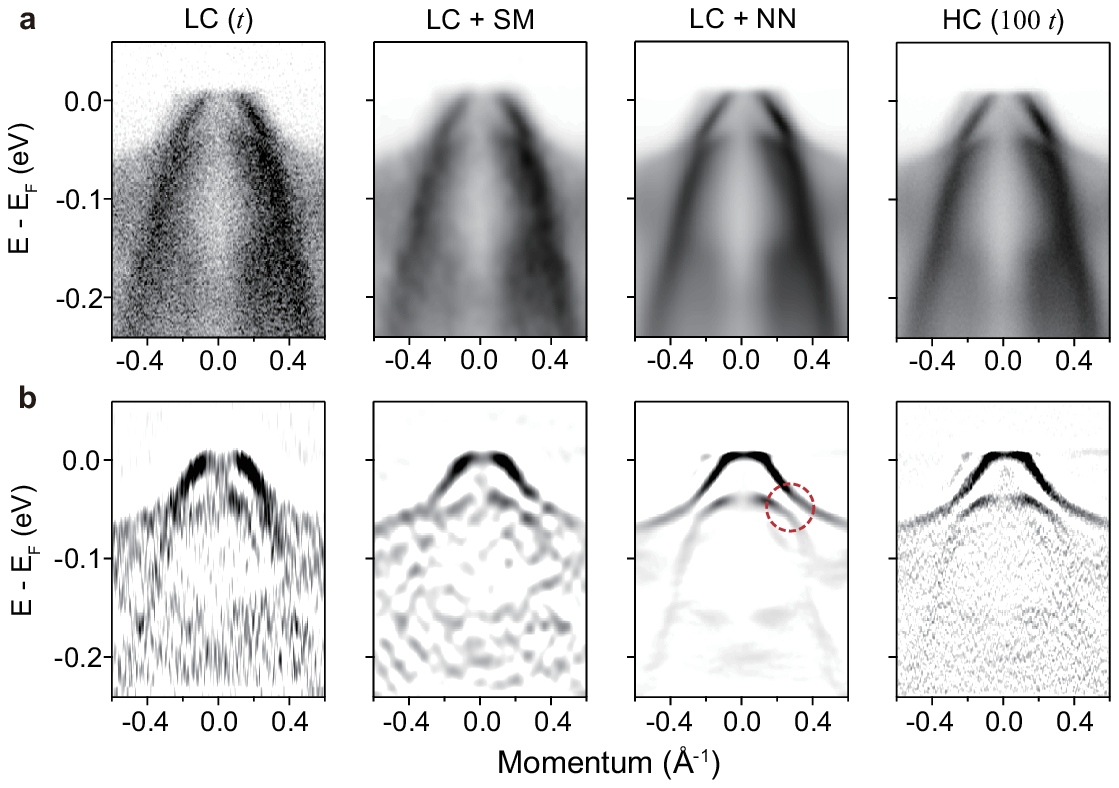}
\caption{Denoising results. (a) FeSe ARPES data long the $M$-$\Gamma$-$M$ cut and denoising results. LC and HC denote low-count and high-count data, respectively. $t$ denotes a unit acquisition time. The low- and high-count data are acquired for $t$ and $100~t$, respectively. SM and NN denotes the Gausian smoothing and denoising neural network, respectively. (b) Corresponding second-derivative results of (a). Red dotted circle represents the hybridization gap between $d_{xy}$ and $d_{xz/yz}$ orbitals.}

\end{figure*}

\section{\label{sec:level1}Introduction}

Advances in a spectroscopic technique often accompany increase in the dimension the experimental technique can cover, allowing for more comprehensive data. An excellent example of this is angle-resolved photoemission spectroscopy (ARPES). A modern-day hemispherical analyzer can take two-dimensional energy/momentum space data whereas a time-of-flight analyzer can cover a three-dimensional energy/momentum space at a time~\cite{damascelli2003angle, damascelli2004probing, zhou2018new, sobota2020electronic}. In addition, the recently developed momentum-resolved photoemission electron microscopy (k-PEEM) can investigate three-dimensional momentum/energy space as well as two-dimensional real space~\cite{sobota2020electronic, medjanik2017direct}. These advances in the analyzer technique allow for new information, and thereby facilitate new discoveries.

An important issue with the aforementioned developments in ARPES is that acquisition of multidimensional data has a fundamental constraint. Owing to the much increased phase space volume to be covered, taking multidimensional data necessarily requires a much longer acquisition time to get the same signal-to-noise ratio (SNR) level~\cite{keunecke2020time}. Increasing the light intensity may not be a solution as, in addition to its own limit, high-intensity light can bring in new issues such as detector non-linearity~\cite{reber2014effects, sobota2020electronic} or space-charge problem~\cite{ishida2016high,hellmann2012vacuum}. Considering the fact that a fresh surface often has a certain lifetime for adequate ARPES~\cite{hofmann2012auger, sobota2020electronic}, this constraint in the acquisition time should be a serious limitation in multidimensional measurements. In fact, the situation also applies to other spectroscopic techniques that acquire multidimensional data. In such a case, the time constraint often affects decision making in real experiments. Therefore, development of a new methodology to overcome the time constraint is highly desired to fully exploit the capability of advanced spectroscopic techniques.

One way to alleviate the limitation is denoising the obtained spectral data. Since the SNR is generally proportional to the square root of the total count, majority of the measurement time is spent on reducing the noise. Thus, if the noise of spectral data can be removed with the intrinsic information preserved, the data acquisition time can be drastically reduced. A conventional way to reduce the noise is the Gaussian smoothing method, exploiting the high-frequency nature of the noise. The Gaussian smoothing is widely used, especially for derivative analyses such as second-derivative or curvature methods, since differentiation highlights the high-frequency signal and is thus vulnerable to the noise~\cite{zhang2011precise}. However, the Gaussian smoothing inevitably blurs data, resulting in loss of the intrinsic information. 

Recent advances in machine learning technology have opened a new era in image processing, especially in removing noises in images. The performance of this new technique is far surpassing the conventional image processing methods, which made the technique widely accepted~\cite{zhang2017beyond, xie2012image}. However, application of the machine learning-based image processing is mostly limited to non-scientific purposes. In light of the remarkable denoising performance of neural networks and the limited data acquisition time in spectroscopic experiments, introduction of the machine learning-based denoising can bring a significant impact on acquisition and analysis of data in spectroscopic experiments. We also note recent studies in which successful application of machine learning has been demonstrated in feature extraction from spectroscopic data~\cite{peng2020super}, self-energy analysis~\cite{yamaji2019hidden}, x-ray structure refinement~\cite{vecsei2019neural, park2017classification, liang2020cryspnet}, and ultrasound spectroscopy~\cite{ghosh2020one}. These examples show that machine learning also can be a useful tool in condensed matter physics.

Here, we demonstrate a deep learning-based denoising method for ARPES data. The proposed method utilizes a deep convolutional neural network to discriminate between noise and intrinsic signals. When the denoising is applied to noisy data for which noise and signal levels are comparable, unlike conventional denoising methods, the intrinsic information seemingly invisible in noisy spectral data is made visible. Our proposed method can drastically reduce the total acquisition time and makes it possible to overcome the limit in the data acquisition time, one of the most serious constraints in spectroscopic experiments. This in turn enables us to fully exploit the advantages of multidimensional measurements. Moreover, the method can be applied to any techniques that acquire multidimensional data and thus suffer from statistical noise due to shorter than desired data acquisition time. \\

\section{\label{sec:level2} Methods}
\subsection{Generation of training data set}

The training data set of the neural network consists of pairs of original and generated data. The neural network is advised to generate high-count data (high SNR) from low-count data (low SNR). The low-count data can be randomly simulated from the high-count data since the count distribution is known to follow the Poisson distribution (see Fig. 1(a))~\cite{he2017visualizing, peng2020super}. The generation of low-count data can be considered as the inverse process of acquisition or denoising since the acquisition or denoising converts the low-count data to high-count data. Thus, only high-count data are required to construct the training data set, which allows us to utilize readily available high-count ARPES data. Furthermore, the proposed random generation method augments the training data set, preventing the neural network from being overfitted. Considering the fact that training of the neural network generally requires numerous data and corresponding labeling, the proposed random generation method is a cost-effective way.

The generation of the training data set is based on the assumption that the count follows the Poisson distribution~\cite{he2017visualizing, peng2020super}. If the total count of data is large enough, the count at a pixel divided by total count converges to the probability of the Poisson distribution.

\begin{equation}\label{eqn1}
\lim_{N \to \infty}{n_{ij} \over N} = P_{ij}
\end{equation}

where $n_{ij}$ denotes counts at pixel $(i,j)$, $N$ denotes the total count in a data and $P_{ij}$ denotes the probability that an electron enters pixel $(i,j)$ when one electron is introduced. Note that $n_{ij}$ and $P_{ij}$ satisfy the following equations, respectively.

\begin{equation}
\sum_{i,j}{n_{ij}} = N,~\sum_{i,j}{P_{ij}} = 1
\end{equation}

The probability can be considered as the intrinsic information that can be obtained from the experiment. If the probability is known, one can randomly simulate experimental data with an arbitrary number of total count. Note that $n_{ij}/N$ is not exactly the same as $P_{ij}$ in ARPES data since the total count of ARPES data is a finite value. We chose the high-count training data which have a sufficiently high total count so that the high-count data have minimum noise and $n_{ij}/N$ is close to $P_{ij}$. Even though the high-count data still has finite noise, the denoising neural network produces noise-free data, since the network is not able to learn to produce the noise due to the random nature of the noise.

The total count in the simulated low-count data ranges from $9\times10^3$ to $3\times10^6$ in (300, 300) grids. Thus, the average count per pixel ranges from 0.1 to 33.3. The wide range of the total counts of the low-count training data sets ensures that the network can denoise data with any statistics. The distribution of the total count in low-count data is set to be log-weight of the total count, so a higher probability is expected for a lower total count data.

\subsection{Training process of the denoising neural network}

The overall training sequence of the neural network is described in Fig. 1(b). From the original high-count data $x$, noisy data $x’$ are generated. Then, the convolutional neural network generates denoised data $f(x’)$ from the noisy input. The size of the training data is set to be $300\times300$. A deep neural network of 20 convolutional layers is adopted to exploit the global contextual information. The structure of the network is based on the network proposed elsewhere~\cite{kim2016accurate}. Each layer of the convolutional neural network has a filter number of 64 and a filter size of 3. After each convolutional layer, the result is passed to a parametric rectifier unit to produce the non-linearity of the network~\cite{he2015delving}. By calculating the loss function $L(x, f(x’))$, the performance of the denoising neural network is determined. The loss function is defined as the weighted sum of mean absolute error and multiscale structural similarity~\cite{wang2003multiscale, zhao2016loss}. Details on the loss function are described in Appendix C. The loss is backpropagated to adjust the parameters used in the neural network~\cite{lecun1989backpropagation}. An Adam optimizer is adopted to train the network for 150 epochs~\cite{kingma2014adam}. The learning rate is initially set to be $5\times10^{-4}$ and multiplied by 0.1 after every 50 epochs for a good convergence. For the training data set, 50 different high-count ARPES data are used and, for each original data, 50 low-count data are randomly generated, resulting in a total of 2500 different low-count data. Note that the FeSe, Bi-2212, Bi$_2$Te$_3$ data in Figs. 2, 3, and 4, respectively, are not included in the training data set. As a data augmentation, the data set is randomly rotated or flipped, and the brightness of the data is also randomly adjusted during the training.

\begin{figure}[htbp]
\includegraphics[width=0.5\textwidth]{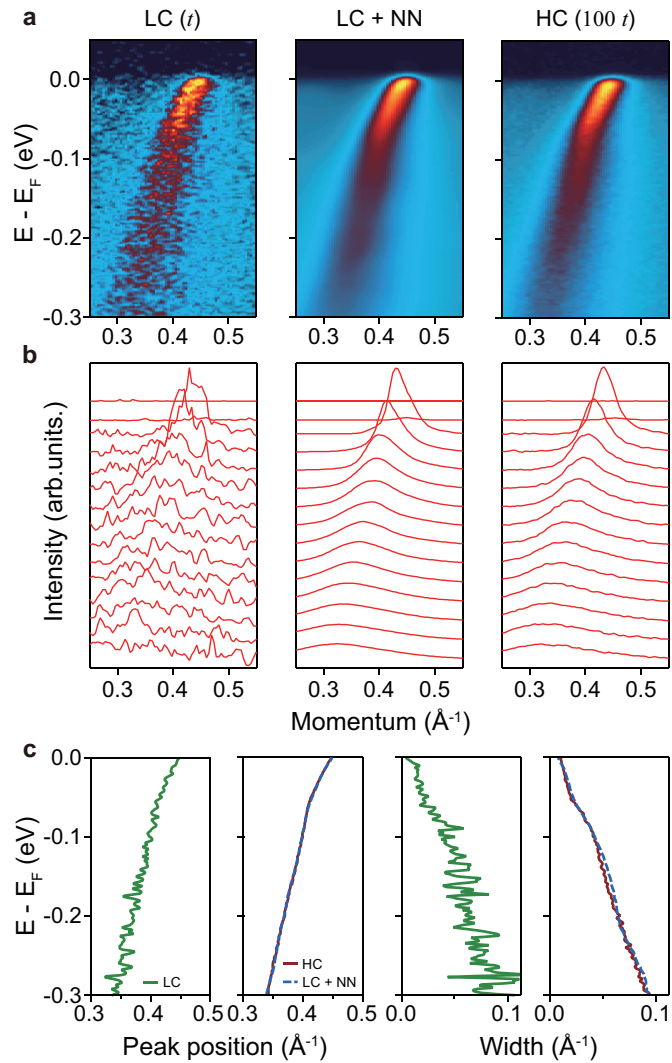}
\caption{Line shape analysis results. (a) Denoising results of ARPES data from Bi-2212 along the nodal cut. (b) Momentum distribution curves (MDCs) of the data in (a). (c)  Peak positions (left) and widths (right) obtained from MDC fitting results of (b).}
\end{figure}

\section{\label{sec:level3}Results}

The result of the denoising neural network is demonstrated in Fig. 2(a). We took ARPES data of FeSe along the $M$-$\Gamma$-$M$ cut for acquisition time $t$ (left) and $100~t$ (right) where $t$ is a unit measurement time. The data acquired for $t$ and $100~t$ are denoted as low-count (LC) and high-count (HC) data, respectively. The LC data is used as the input to the denoising neural network, and the HC data is compared with the denoised data. The LC data shows a high level of noise due to the low total count. The network produces noise-free data (middle panel), even though the input data is quite noisy. The LC data after the denoising neural network (LC + NN) is comparable to the HC data, showing almost the same features. Yet, we note that small features of the denoised data are a bit blurry due to the lack of information in the LC data. To visualize the band structure more clearly, we plot in Fig. 2(b) the second derivative of the data. Band dispersions from the LC data are barely visible. On the other hand, the second derivative of the denoised data in the middle panel shows very clear band dispersions, especially the hybridization gap between $d_{xy}$ and $d_{xz/yz}$ orbitals at $\pm$0.3~\AA$^{-1}$ as indicated by a red dotted circle~\cite{huh2020absence, zhang2015observation} which is not resolved in the original LC data. Note that, since the noise is removed after the denoising neural network, the second derivative of the denoised data shows clean spectra whereas the HC data has residual noise despite the long acquisition time.

To verify the validity of the denoising neural network in a quantitative way, we conducted line shape analysis on denoised Bi-2212 data taken along the nodal cut. As can be seen in Fig. 3(a), the denoising neural network preserves intrinsic band structure while removes the noise. Removal of the noise is more clearly seen in the momentum distributions curves (MDCs) of the data depicted in Fig. 3(b). Line shape analysis was conducted by fitting the MDCs to obtain the peak position and width as depicted in Fig. 3(c). The fitting results of LC + NN and HC data are almost identical, directly demonstrating that the denoising neural network preserves the quantitative information of the band structure. We wish to point out that the well-known 70~meV kink at the nodal point is clearly resolved for both LC + NN and HC data~\cite{garcia2010through, kondo2013anomalous}, whereas the fitting result of the raw LC data is too noisy to identify the kink position.

\begin{figure}[htbp]
\includegraphics[width=0.5\textwidth]{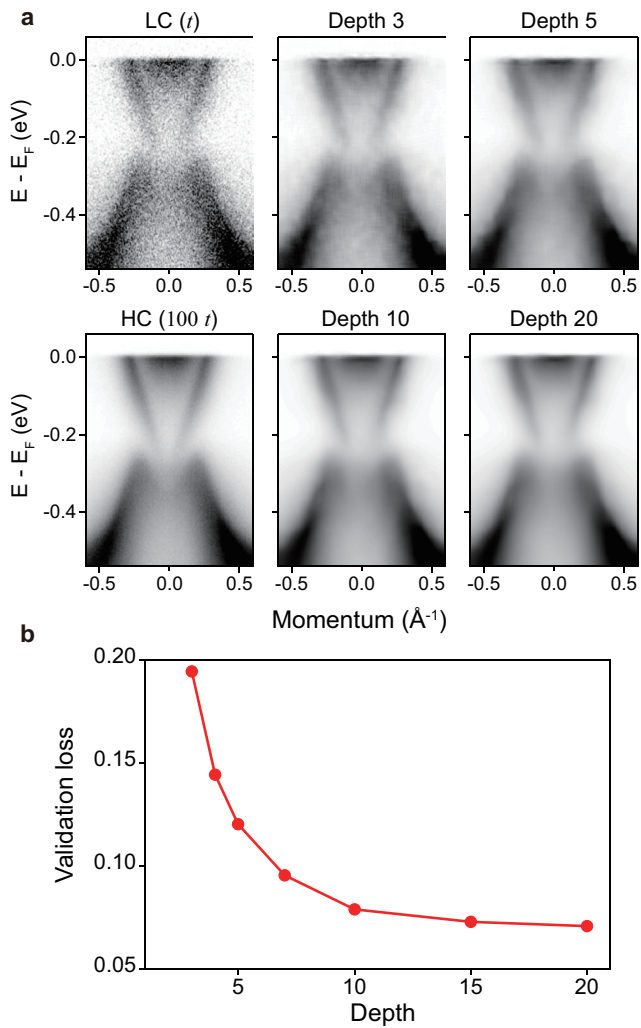}
\caption{Depth dependent denoising results. (a) ARPES data of Bi$_2$Te$_3$ along the $K$-$\Gamma$-$K$ cut and depth dependent denoising results. Depth denotes the number of convolutional layers in the denoising neural network. (b) Validation loss $L(x, f(x’))$ as a function of the depth.}
\end{figure}

\section{\label{sec:level4}Discussion}

\subsection{Understanding effectiveness of denoising neural network}

The reason why the deep learning-based denoising is effective for ARPES data may be summarized into one sentence; the neighboring pixel values in ARPES data are correlated with each other. Two major factors contribute to the correlation. Firstly, the typical dimension for ARPES features is larger than the data pixel size, leading to occupation of several pixels for any feature. Thus, if the value at a pixel is large, it is likely for neighboring pixels to have a large value. Secondly, the length scale over which the band structure changes is larger than the data pixel size. Hence, the band structure does not change abruptly over the length scale of the pixel. This means that the band structure has an approximate translational symmetry in a short length scale. Even if the information at a pixel is corrupted with noise, the value at the pixel can be recovered from the most statistically probable value inferred from adjacent pixel values. Therefore, a data set carries more information than just the pixel-wise sum of information. With the additional information, the seemingly imperfect information of the noisy data can be recovered.

In order to extract such kind of contextual information, the neural network should accept global information of data. Since the receptive field of a convolutional neural network is $(2 D + 1)\times(2 D + 1)$ for a depth $D$ network with a filter size of 3, a deeper network of a larger $D$ receives more global information~\cite{kim2016accurate,he2016deep}. Here, depth means the number of convolutional layers in a neural network. For a large receptive area, we adopted a convolutional neural network of 20 layers. We experimentally show in Fig. 4 that a deeper network tends to work better. We took LC and HC ARPES data of Bi$_2$Te$_3$ thin film grown on Si(111) substrate along the $K$-$\Gamma$-$K$ cut (see Fig. 4(a)). The LC data is then passed to the denoising neural network with different numbers of convolutional layers. It is seen that, as the depth of the denoising neural network increases, the noise is better removed and the denoised data become more similar to the HC data. In Fig. 4(b), the validation loss $L(x, f(x’))$ is plotted as a function of the depth to visualize the tendency more clearly. The validation data set consists of 20 pairs of LC and HC data obtained from ARPES measurements. The validation loss monotonically decreases with increasing depth. Generally, a deeper network tends to work better if the network is not overfitted.~\cite{he2016deep} We note that the network deeper than 20~layers could not be stably trained due to gradient vanishing/exploding. Further studies are needed to train a deeper network.

Considering the mechanism of the denoising neural network, it is expected that the network does not work well for data which are very different from the training data. For instance, we found that denoising performance for data with a large background was not very good, since very few data sets with high-level of background were included in the training data set. This point may be improved by including more training data with a variety of features.

\subsection{Application to higher-dimensional data}

Finally, we discuss possible application of the denoising neural network to higher-dimensional data. Since the basic principle of denoising is the inference from adjacent pixel values, better denoising performance is expected if the data have more neighboring pixels. That is, there is much more contextual information that can be extracted from neighboring pixels in higher-dimensional data and the denoising neural network should work better. For the same reason, the denoising neural network may not work well for one-dimensional spectra due to relatively small number of neighboring pixels. Considering the fact that data acquisition in a multidimensional phase space takes a long time, the denoising neural network will be a method to alleviate the time constraint, thereby allowing us to fully exploit the advantages of multidimensional measurements, not only for ARPES but possibly for other time demanding experimental techniques.

\section{\label{sec:level5}Summary and outlook}


We have demonstrated that the network not only reduces the acquisition time but makes the noise of data to an experimentally unreachable level. If trained properly, this scheme can be also used for distributions other than Poisson. The thorough removal of the noise opens up a new route for data analysis techniques for which noise is an apparent obstacle such as deconvolution~\cite{rameau2010application} and self-energy analysis~\cite{kordyuk2005bare}. That is, the denoising neural network can be used as the base layer for other data analysis techniques, which calls for further studies on artificial intelligence-based data analysis methods.

It is also noteworthy that open-source Python-based data analysis packages for ARPES and other multidimensional experimental techniques were recently demonstrated.~\cite{kramer2021visualization, stansbury2020pyarpes} Considering the open-source nature of these packages along with the high expandability of the Python, implementation of our deep learning-based denoising method into one of the Python-based data analysis packages can be easily achieved. In addition, since the time required to denoise an ARPES data set typically takes less than a few seconds, the implementation will allow us to denoise data in real-time, while analyzing and visualizing data. This would provide a new deep learning-based data analysis platform for the ARPES community.

\section*{Supplementary material}
See supplementary material for the demonstrative denoising program that can be accessed via a webpage.

\section*{Acknowledgements}
We are grateful to Y. Ishida for valuable discussion and useful comments. This work was supported by the Institute for Basic Science in Korea (Grant No. IBS-R009-G2).

\appendix

\section{Training data set}

The data used for training are plotted in Fig. 5. Data complying with the following conditions were used as training data. Firstly, the data must have a sufficiently high total count to have a low level of noise. Secondly, the data has no artifacts from detector inhomogeneity.

\begin{figure}[htbp]
\includegraphics[width=0.5\textwidth]{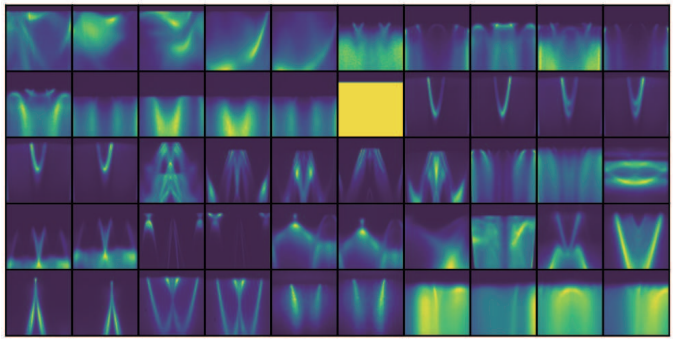}
\caption{Plot of data set used in the training}
\end{figure}

\begin{figure}[htbp]
\includegraphics[width=0.5\textwidth]{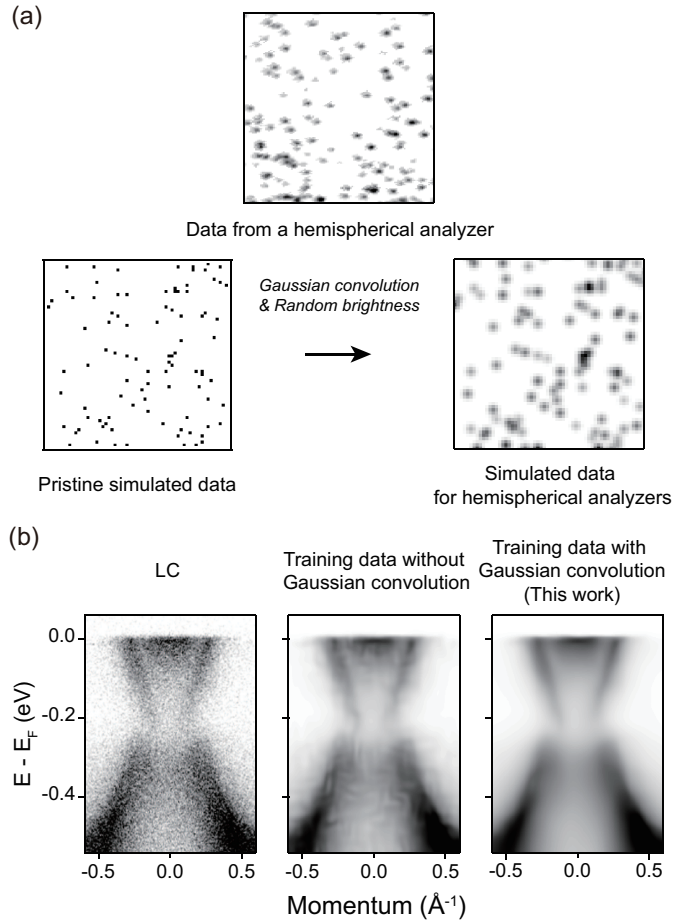}
\caption{(a) Examples of the data from a hemispherical analyzer and simulated data. (b) ARPES data of Bi$_2$Te$_3$ and training data dependent denoising results.}
\end{figure}

\section{Low-count data for hemispherical analyzers}

The low-count data generated from the aforementioned method has pixel-wise discrete values. However, the data acquired from a hemispherical analyzer consists of counts occupying several pixels as described in Fig. 6(a) since the detector of a hemispherical analyzer measures an impinged signal on a phosphor screen using a CCD camera~\cite{sobota2020electronic}. To simulate the experimental results obtained with a hemispherical analyzer, the simulated data is convoluted with a Gaussian function with random peak intensities and widths. Without such Gaussian convolution, denoising the data from a hemispherical analyzer shows bad performance (See Fig. 6(b)). 

\begin{figure}[htbp]
\includegraphics[width=0.5\textwidth]{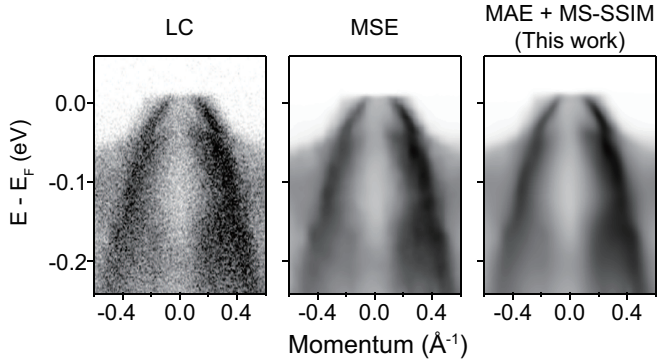}
\caption{ARPES data of FeSe and loss function dependent denoising results. MSE denotes mean squared error and MS-SSIM denotes multiscale structural similarity.}
\end{figure}

\section{Loss function}

The loss function adopted in this work is a weighted sum of mean absolute error (MAE) and multiscale structural similarity (MS-SSIM). The conventional mean squared error (MSE) has a weak penalty for a small difference. Hence, the denoised result is blurry since making the data blurry is an easy way to minimize MSE loss\cite{zhao2016loss}. We therefore adopted a new loss function consists of MAE and MS-SSIM as described elsewhere~\cite{zhao2016loss}. 

\begin{equation}\label{eqn2}
L = (1-\alpha) \cdot L_{MAE} + \alpha \cdot L_{MS-SSIM}
\end{equation}

where $\alpha$ is set to be 0.7. The MAE loss has a higher penalty for a small difference compared to MSE loss so the result is expected to be less blurry. The MS-SSIM loss catches a similarity over a wide range of the data compared to MSE loss or MAE loss which calculates the pixel-wise difference. Thus, the result is perceptually more plausible and the MS-SSIM loss ensures overall similarity. The comparison results of MSE loss and the loss used in this work are plotted in Fig. 7.

\begin{figure}[htbp]
\includegraphics[width=0.5\textwidth]{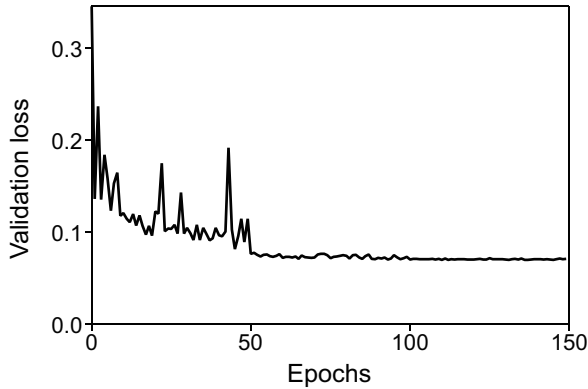}
\caption{Plot of validation loss during the training.}
\end{figure}

\section{Overfitting}

Overfitting is one of the most serious issues in deep learning. Generally, the overfitting occurs when the training data set is small compared to the size of the neural network. We checked the overfitting by monitoring validation loss during the training since the increase of the validation loss is a representative symptom of the overfitting. As shown in Fig. 8, the validation loss converges to a value at the end of the training. From the result, we judged that the model is not overfitted.

\section*{Data availability}
The data that support the findings of this study are available from the corresponding author upon reasonable request.

\nocite{*}
\bibliography{references}

\end{document}